\documentclass{article}

\usepackage{microtype}
\usepackage{graphicx}
\usepackage{subcaption}
\usepackage{booktabs} 
\usepackage{multirow}

\usepackage{hyperref}

\usepackage[preprint]{icml2026}

\usepackage{amsmath}
\usepackage{amssymb}
\usepackage{mathtools}
\usepackage{amsthm}

\usepackage[capitalize,noabbrev]{cleveref}

\theoremstyle{plain}

\theoremstyle{definition}

\theoremstyle{remark}

\usepackage[textsize=tiny]{todonotes}

\icmltitlerunning{OptiMAG: Structure-Semantic Alignment via Unbalanced Optimal Transport}

\begin{document}

\twocolumn[
\icmltitle{OptiMAG: Structure-Semantic Alignment via Unbalanced Optimal Transport}

\icmlsetsymbol{equal}{*}

\begin{icmlauthorlist}
\icmlauthor{Yilong Zuo}{bit}
\icmlauthor{Xunkai Li}{bit}
\icmlauthor{Zhihan Zhang}{bit}
\icmlauthor{Qiangqiang Dai}{bit}
\icmlauthor{Ronghua Li}{bit}
\icmlauthor{Guoren Wang}{bit}
\end{icmlauthorlist}

\icmlaffiliation{bit}{Beijing Institute of Technology, Beijing, China}

\icmlcorrespondingauthor{Ronghua Li}{lironghuabit@126.com}

\icmlkeywords{Multimodal Learning, Multimodal Attributed Graphs, Optimal Transport, Graph Neural Networks, Gromov-Wasserstein}

\vskip 0.3in
]

\printAffiliationsAndNotice{} 

\begin{abstract}
Multimodal Attributed Graphs (MAGs) have been widely adopted for modeling complex systems by integrating multi-modal information, such as text and images, on nodes. However, we identify a discrepancy between the implicit semantic structure induced by different modality embeddings and the explicit graph structure. For instance, neighbors in the explicit graph structure may be close in one modality but distant in another. Since existing methods typically perform message passing over the fixed explicit graph structure, they inadvertently aggregate dissimilar features, introducing modality-specific noise and impeding effective node representation learning. To address this, we propose \textbf{OptiMAG}, an Unbalanced Optimal Transport-based regularization framework. OptiMAG employs the Fused Gromov-Wasserstein distance to explicitly guide cross-modal structural consistency within local neighborhoods, effectively mitigating structural-semantic conflicts. Moreover, a KL divergence penalty enables adaptive handling of cross-modal inconsistencies. This framework can be seamlessly integrated into existing multimodal graph models, acting as an effective drop-in regularizer. Experiments demonstrate that OptiMAG consistently outperforms baselines across multiple tasks, ranging from graph-centric tasks (e.g., node classification, link prediction) to multimodal-centric generation tasks (e.g., graph2text, graph2image). The source code will be available upon acceptance.
\end{abstract}

\begin{figure}[htbp]
    \centering
    \includegraphics[width=0.4\textwidth]{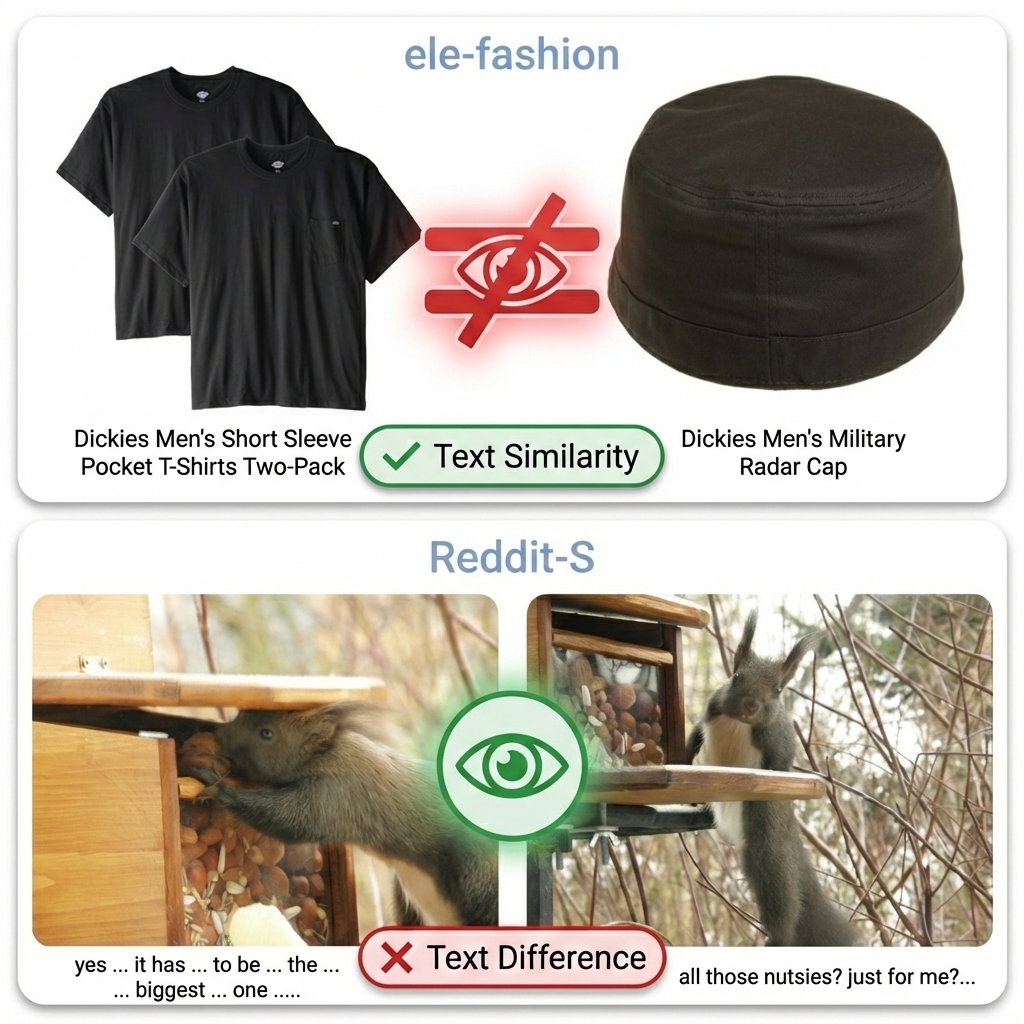}
    \caption{Cross-modal inconsistency in real MAG edges. \textbf{Top} (ele-fashion): two connected product nodes share the ``Dickies Men's'' brand (textually similar) yet differ visually (T-shirt vs.\ cap). \textbf{Bottom} (Reddit-S): two connected posts show similar squirrel-with-nut imagery but diverge in text content.}
    \label{fig:intro_example}
\end{figure}

\section{Introduction}
\label{sec:intro}


Multimodal Attributed Graphs (MAGs), which integrate graph topology with rich multimodal node features, have gained traction in various domains including e-commerce recommendation and social network analysis \cite{yanWhenGraphMeets2024, zhuMosaicModalitiesComprehensive2025}. Unlike Text-Attributed Graphs (TAGs) that carry only textual features, each node in a MAG is associated with multiple modalities (e.g., text and images). This enriches node semantics yet introduces challenges in learning coherent representations from such heterogeneous inputs.

Prevailing approaches \cite{fangGRAPHGPTOSynergisticMultimodal2025, heUniGraph2LearningUnified2025} usually follow a two-stage pipeline: (\textit{i}) project features of each modality into a shared embedding space and optionally fusing them; (\textit{ii}) propagate information from neighbors via Graph Neural Networks (GNNs). This paradigm implicitly assumes high \textit{homophily}, i.e., that adjacent nodes share similar features for all modalities.

However, this assumption often breaks down in MAGs. As \cref{fig:intro_example} 
shows, two adjacent nodes can be highly similar in one modality yet dissimilar 
in another. This cross-modal inconsistency has been theoretically characterized through selection and perturbation biases in multimodal representation learning \cite{caiValueCrossModalMisalignment2025}. Our empirical study across multiple MAG datasets 
(\cref{appendix:empirical_analysis}) also confirms that such cross-modal 
inconsistency is pervasive: edges frequently connect nodes that share features 
in only a \textit{subset} of modalities. We attribute this to the nature of MAG 
edge formation: an edge typically reflects similarity in \textit{some} modality, 
without requiring agreement across \textit{all} modalities. 

To characterize this phenomenon, we distinguish two topologies: the 
\textit{implicit semantic structure}, an ideal graph where edges link only 
feature-similar nodes within a given modality, and the observed 
\textit{explicit graph structure}, which is shaped by multi-modal factors 
and thus includes edges that are homophilous for some modalities but 
heterophilous for others. We term this mismatch \textbf{structural-semantic conflict}.

This conflict affects GNN message passing in nuanced ways. On one hand, heterophilous edges can propagate features that are inconsistent within a given modality, injecting modality-specific noise and degrading learned representations. On the other hand, not every such edge is detrimental: when two nodes are semantically related yet visually distinct (e.g., ``book'' and ``bookshelf''), aggregating their divergent visual features can enrich the representation by capturing complementary cues \cite{liLearnHeterophily2024}.

The challenge is that current multimodal graph models \cite{huMMGCNMultimodalFusion2021,taoMGATMultimodalGraph2020} apply uniform aggregation, conflating beneficial complementarity with harmful conflict and thereby admitting destructive noise. Meanwhile, heterophilous-graph techniques---including spectral filtering \cite{zhuBeyondHomophilyGraph2020}, graph rewiring \cite{liDeepHeterophilyGraph2024}, and non-local aggregation (cf.~\cite{zhengGNNHeterophilySurvey2024} for a survey)---target unimodal settings; naively decoupling modalities forfeits cross-modal guidance and integrates poorly with existing pipelines. What is needed, therefore, is an alignment mechanism that (i) bridges each modality's implicit semantic structure with the explicit graph, (ii) distinguishes beneficial complementarity from harmful conflict rather than enforcing rigid alignment, and (iii) operates as a plug-and-play regularizer for seamless integration with state-of-the-art multimodal graph models. We term this goal \textbf{structural-semantic alignment}.

To meet these requirements, we present \textbf{OptiMAG} (\cref{fig:framework}), 
a regularization framework based on Unbalanced Optimal Transport (UOT). 
OptiMAG recasts structural-semantic alignment as a distributional transport 
problem and employs a hybrid strategy: the Fused Gromov-Wasserstein (FGW) 
distance \cite{vayer2019optimal} aligns cross-domain structures (addressing~\textit{i}), while a 
KL divergence penalty enables adaptive rejection of harmful edges 
(addressing~\textit{ii}). As a drop-in regularizer (addressing~\textit{iii}), 
OptiMAG integrates seamlessly with existing backbones. Because computing GW 
scales as $O(N^3)$, we accelerate it via subgraph sampling and 
entropic Sinkhorn iterations, making OptiMAG practical for large-scale graphs.

We evaluate OptiMAG on six MAG benchmarks covering both graph-centric 
(node classification, link prediction, clustering) and generation tasks 
(graph-to-text, graph-to-image). OptiMAG yields consistent gains across 
diverse backbones, improving node classification accuracy by up to 4.6\% 
and captioning CIDEr by over 4 points. The improvement is most pronounced 
for pretrained encoders like UniGraph2, where OptiMAG's structured 
regularization steers large models toward finer-grained graph adaptation.

\textbf{Our Contributions.} (1) \textit{Problem Formalization.} We formalize \textit{structural-semantic conflict} and provide empirical evidence for its prevalence in MAGs and its adverse effect on GNN aggregation. (2) \textit{Method.} We introduce OptiMAG, an Unbalanced Optimal Transport-based regularizer that mitigates aggregation noise stemming from cross-modal structural misalignment. (3) \textit{Validation.} We validate OptiMAG on diverse backbones across graph-centric (node classification, link prediction, clustering) and multimodal-centric (graph-to-text, graph-to-image) tasks, showing consistent gains over strong baselines.

\section{Preliminaries}
\label{sec:preliminaries}

\subsection{Multimodal Attributed Graphs}
\label{subsec:mag_def}

A Multimodal Attributed Graph is defined as $\mathcal{G} = (\mathcal{V}, \mathcal{E}, \mathcal{M})$~\cite{yanWhenGraphMeets2024, zhuMosaicModalitiesComprehensive2025}, where $\mathcal{V}$ is a set of $N$ nodes, $\mathcal{E}$ the edge set, and $\mathcal{M} = \{T, I\}$ the modality set (text and image).

For each node $v_i \in \mathcal{V}$, let $\mathbf{x}_i^T$ and $\mathbf{x}_i^I$ denote the raw textual and visual inputs. Modality-specific encoders $f_{\theta_T}$ and $f_{\theta_I}$ map these inputs to latent representations:
\begin{equation}
    \mathbf{h}_i^T = f_{\theta_T}(\mathbf{x}_i^T) \in \mathbb{R}^{d_T}, \quad \mathbf{h}_i^I = f_{\theta_I}(\mathbf{x}_i^I) \in \mathbb{R}^{d_I}
\end{equation}
where $d_T$ and $d_I$ are the respective embedding dimensions. Typically $d_T \neq d_I$, so the two modality spaces are not naturally aligned.

\subsection{Optimal Transport}

Optimal Transport (OT) formalizes the comparison of probability distributions as a constrained mass transfer problem~\cite{peyre2019computational}. Given two discrete distributions $\boldsymbol{\mu} \in \Sigma_n$ and $\boldsymbol{\nu} \in \Sigma_m$ over source domain $\mathcal{X}$ and target domain $\mathcal{Y}$, where $\Sigma_n = \{ \mathbf{a} \in \mathbb{R}_+^n | \sum_i a_i = 1 \}$ denotes the probability simplex, the Kantorovich formulation seeks a transport plan $\boldsymbol{\pi} \in \mathbb{R}_+^{n \times m}$ that minimizes the total cost of moving mass from $\boldsymbol{\mu}$ to $\boldsymbol{\nu}$:

\begin{equation}
\begin{aligned}
    \mathcal{W}(\boldsymbol{\mu}, \boldsymbol{\nu}) &= \min_{\boldsymbol{\pi} \in \Pi(\boldsymbol{\mu}, \boldsymbol{\nu})} \sum_{i=1}^n \sum_{j=1}^m \mathbf{M}_{ij} \boldsymbol{\pi}_{ij} \\
    &= \min_{\boldsymbol{\pi} \in \Pi(\boldsymbol{\mu}, \boldsymbol{\nu})} \langle \boldsymbol{\pi}, \mathbf{M} \rangle_F
\end{aligned}
\end{equation}
Here, $\Pi(\boldsymbol{\mu}, \boldsymbol{\nu}) = \{ \boldsymbol{\pi} \in \mathbb{R}_+^{n \times m} | \boldsymbol{\pi}\mathbf{1}_m = \boldsymbol{\mu}, \boldsymbol{\pi}^\top\mathbf{1}_n = \boldsymbol{\nu} \}$ defines the set of couplings satisfying marginal constraints. The ground cost matrix $\mathbf{M} \in \mathbb{R}_+^{n \times m}$ encodes pairwise transport costs, with $\mathbf{M}_{ij} = \|x_i - y_j\|^2$ being a common choice.

When $\mathbf{M}$ equals the $p$-th power of a metric, the optimum defines the $p$-Wasserstein distance, widely used for measuring distributional discrepancy. Entropic regularization~\cite{cuturi2013sinkhorn} is often applied to speed up computation and obtain differentiable gradients.

\subsection{Gromov-Wasserstein Distance}

The standard Wasserstein distance assumes both domains share the same metric space for computing $\mathbf{M}_{ij}$. In multimodal graph learning, the source and target often reside in entirely different metric spaces---semantic embeddings versus graph topology, for instance---rendering direct Wasserstein comparison infeasible.

The Gromov-Wasserstein (GW) distance~\cite{memoli2011gromov} sidesteps this issue: instead of comparing absolute point-to-point distances, it matches \textit{intra-domain relations}. Given two spaces $(\mathcal{X}, \mathbf{C}_{\mathcal{X}})$ and $(\mathcal{Y}, \mathbf{C}_{\mathcal{Y}})$ equipped with structural cost matrices, GW is defined as:
\begin{equation}
    \mathcal{GW}(\boldsymbol{\mu}, \boldsymbol{\nu}) = \min_{\boldsymbol{\pi} \in \Pi(\boldsymbol{\mu}, \boldsymbol{\nu})} \sum_{i,j,k,l} L(\mathbf{C}_{\mathcal{X}}(i,k), \mathbf{C}_{\mathcal{Y}}(j,l)) \boldsymbol{\pi}_{ij} \boldsymbol{\pi}_{kl}
\end{equation}
where $L(a, b) = |a - b|^2$ penalizes relational discrepancies. In effect, GW seeks a coupling under which relative distances are preserved across domains---a property that suits heterogeneous topology alignment. To jointly exploit feature similarity and structural correspondence, Fused Gromov-Wasserstein (FGW)~\cite{vayer2019optimal,vayer2019fused} combines Wasserstein and GW objectives. Our framework extends FGW to the \emph{unbalanced} setting~\cite{sejourne2019sinkhorn}, enabling adaptive mass re-weighting to handle structural-semantic conflicts.

\begin{figure*}[t]
    \centering
    \includegraphics[width=\textwidth]{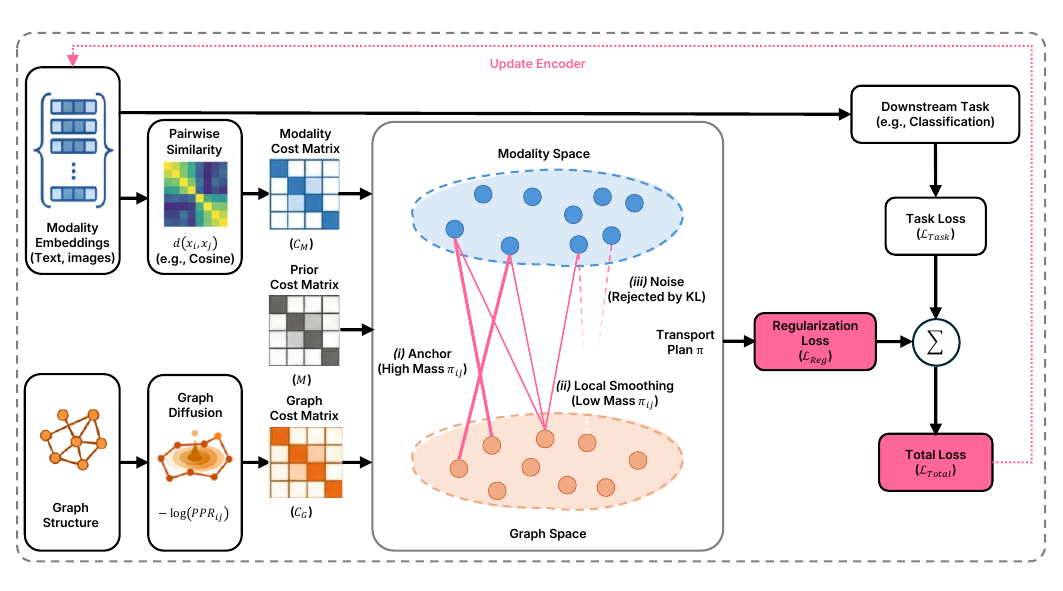}
    \caption{Overview of OptiMAG's Unbalanced Fused Gromov-Wasserstein (UFGW) alignment. \textbf{Left:} For each batch, we build a modality cost matrix $C_M$ (cosine distances) and a graph cost matrix $C_G$ (PPR-based diffusion distances), along with a prior alignment matrix $M$. \textbf{Center:} Sinkhorn iterations yield a transport plan $\pi$ that exhibits three behaviors—(i) \textit{Anchor}: high-consistency nodes retain large mass; (ii) \textit{Local Smoothing}: mildly misaligned nodes spread mass to neighbors; (iii) \textit{Noise Rejection}: severely conflicting nodes receive negligible mass via KL penalty. \textbf{Right:} The regularization loss $\mathcal{L}_{\mathrm{Reg}}$ is combined with the task loss $\mathcal{L}_{\mathrm{Task}}$ to update the encoder.}
    \label{fig:framework}
\end{figure*}

\section{The OptiMAG Framework}
\label{sec:method}

OptiMAG is a plug-and-play regularization framework that mitigates structural-semantic conflicts by aligning each modality's implicit semantic topology with the explicit graph geometry. As shown in Figure~\ref{fig:framework}, it proceeds in three stages: (\textit{i}) mapping heterogeneous modality features and the graph structure into unified metric measure spaces; (\textit{ii}) learning a transport plan via the Unbalanced Fused Gromov-Wasserstein (UFGW) distance with adaptive denoising; and (\textit{iii}) using the transport cost as a regularization term to guide encoder learning.

\subsection{Geometric Space Construction}
\label{subsec:space_construction}

Our formulation treats both modality embeddings and the graph structure as Metric Measure Spaces~\cite{peyre2019computational}, a standard framework in optimal transport theory. Concretely, given a sampled batch $\mathcal{B}$ of $N$ nodes, we represent each domain as a triplet $(\boldsymbol{\mu}, \mathcal{X}, \mathbf{C})$, where $\boldsymbol{\mu}$ is a probability distribution over nodes, $\mathcal{X}$ the underlying feature space, and $\mathbf{C} \in \mathbb{R}^{N \times N}$ a cost matrix that encodes pairwise geometric relations within the domain.

\subsubsection{Semantic Modality Space}
Consider a modality $m \in \{T, I\}$. A modality-specific encoder $f_{\theta_m}$---BERT for text or ResNet for images, in our implementation---first maps raw inputs $\mathbf{X}^m_{\mathcal{B}}$ to dense embeddings $\mathbf{H}^m = f_{\theta_m}(\mathbf{X}^m_{\mathcal{B}}) \in \mathbb{R}^{N \times d_m}$.

The pairwise semantic structure is then captured through cosine distance:
\begin{equation}
    \mathbf{C}_m(i, j) = 1 - \frac{\mathbf{h}^m_i \cdot \mathbf{h}^m_j}{\|\mathbf{h}^m_i\|_2 \|\mathbf{h}^m_j\|_2}
\end{equation}
We prefer cosine over Euclidean distance because the direction of semantic vectors---rather than magnitude---tends to better capture semantic similarity in pretrained language and vision models~\cite{reimers2019sentence}. A smaller $\mathbf{C}_m(i, j)$ thus reflects a stronger semantic affinity between nodes $i$ and $j$.

Since the embedding scales of different modalities can vary substantially---and the graph-based cost matrix introduced later differs further still---we normalize each cost matrix by its mean:
\begin{equation}
    \bar{\mathbf{C}}_m = \frac{\mathbf{C}_m}{\text{mean}(\mathbf{C}_m) + \delta}
\end{equation}
where $\delta$ is a small constant for numerical stability. For the node distribution, we adopt a uniform prior $\boldsymbol{\mu}_m = \frac{1}{N}\mathbf{1}_N$, treating all nodes as equally important a priori.

\subsubsection{Explicit Graph Space}
Unlike the continuous modality embeddings, the graph structure is inherently discrete: adjacency matrix $\mathbf{A}$ is sparse and encodes only direct connections. Directly using $\mathbf{A}$ as a cost matrix would lose higher-order proximity and impede gradient flow. We therefore smooth the topology via Personalized PageRank (PPR) diffusion~\cite{gasteiger2019predict}, a technique widely adopted in GNNs for capturing higher-order proximity. Let $\tilde{\mathbf{A}} = \mathbf{D}^{-1}\mathbf{A}$ denote the row-normalized random walk matrix; we precompute
\begin{equation}
    \mathbf{\Pi} = \beta (\mathbf{I} - (1-\beta)\tilde{\mathbf{A}})^{-1}
\end{equation}
over the full graph, where $\beta$ is the teleport probability. At training time, each batch $\mathcal{B}$ simply indexes the relevant submatrix $\mathbf{S}_{\mathcal{B}} = \mathbf{\Pi}[\mathcal{B}, \mathcal{B}]$, retaining global structural context without recomputation.

Finally, we convert the diffusion probabilities into a proper distance by taking the negative log:
\begin{equation}
    \mathbf{C}_G(i, j) = -\log(\mathbf{S}_{\mathcal{B}}(i, j) + \delta)
\end{equation}
This log-transform turns multiplicative proximity into additive distance, making the graph geometry comparable to the cosine-based modality geometry. The same mean-normalization yields $\bar{\mathbf{C}}_G$, and we again use a uniform node distribution $\boldsymbol{\nu} = \frac{1}{N}\mathbf{1}_N$.

\subsection{Unbalanced Fused Alignment Mechanism}
\label{subsec:ufgw_mechanism}

Having cast both domains as MMS, we now turn to the alignment objective itself. A naive approach would enforce rigid one-to-one matching, but this ignores two realities: first, structurally adjacent nodes may exhibit complementary---rather than identical---semantics; second, some edges are inherently noisy. OptiMAG addresses these issues through an Unbalanced Fused Gromov-Wasserstein (UFGW) formulation~\cite{vayer2019optimal} that couples node identity with structural consistency while allowing selective rejection of outliers.

\subsubsection{Fused Transport Costs}
Our transport cost has two components: a \textit{linear} term that anchors node identities, and a \textit{quadratic} term that enforces relational consistency.

\textbf{Linear term (anchor prior).} In our setting, node $i$ in the modality space inherently corresponds to node $i$ in the graph---they represent the same entity. We encode this prior through an anchor matrix $\mathbf{M} \in \mathbb{R}^{N \times N}$:
\begin{equation}
    \mathbf{M}_{ij} = \begin{cases} 
    0 & \text{if } i = j \\ 
    \tau & \text{if } i \neq j 
    \end{cases}
\end{equation}
Rather than setting off-diagonal entries to infinity (which would enforce hard matching), we use a finite threshold $\tau > 0$. This design encourages diagonal transport when structures align well (Anchor in Figure~\ref{fig:framework}), yet allows mass to spread to neighbors when doing so reduces overall structural discrepancy (Local Smoothing).

\textbf{Quadratic term (structural consistency).} The linear term alone does not capture whether pairwise relationships are preserved. To this end, we borrow from Gromov-Wasserstein and penalize relational mismatches:
\begin{equation}
    L(\bar{\mathbf{C}}_m(i, k), \bar{\mathbf{C}}_G(j, l)) = | \bar{\mathbf{C}}_m(i, k) - \bar{\mathbf{C}}_G(j, l) |^2
\end{equation}
The idea is that transport should preserve relative geometry: if the plan maps $i \to j$ and $k \to l$, then the modality-space distance $\bar{\mathbf{C}}_m(i,k)$ should match the graph-space distance $\bar{\mathbf{C}}_G(j,l)$. Any discrepancy incurs a quadratic penalty, discouraging plans that distort local structure.

\subsubsection{Mass Relaxation via KL Divergence}
Classical OT requires the transport plan $\boldsymbol{\pi}$ to satisfy strict marginal constraints: $\boldsymbol{\pi}\mathbf{1}_N = \boldsymbol{\mu}_m$ and $\boldsymbol{\pi}^\top\mathbf{1}_N = \boldsymbol{\nu}$. In practice, this is problematic. When some edges are inherently noisy, forcing complete mass transport creates spurious correspondences that pollute encoder gradients. We instead adopt the Unbalanced OT framework~\cite{chizat2018scaling}, relaxing these hard constraints into soft penalties via KL divergence:
\begin{equation}
    \mathcal{L}_{\text{KL}} = \rho \left( \text{KL}(\boldsymbol{\pi}\mathbf{1}_N \| \boldsymbol{\mu}_m) + \text{KL}(\boldsymbol{\pi}^\top\mathbf{1}_N \| \boldsymbol{\nu}) \right)
\end{equation}

\textbf{How KL penalty distinguishes complementarity from conflict.}
The crux of this formulation lies in the trade-off each node faces between alignment cost and marginal fidelity. Let $\boldsymbol{r}_i = \sum_j \boldsymbol{\pi}_{ij}$ denote the row marginal, i.e., the total mass shipped from node $i$. For a node suffering severe structural-semantic conflict, shipping its full quota $\mu_i$ incurs a large alignment cost. Alternatively, the optimizer can shrink $\boldsymbol{r}_i$ below $\mu_i$, paying a KL penalty $\rho \cdot \mu_i \log(\mu_i / \boldsymbol{r}_i)$ instead.

Balancing these two forces yields an equilibrium:
\begin{equation}
    \boldsymbol{r}_i^* \propto \mu_i \exp\left( -\frac{c_i}{\rho} \right)
    \label{eq:mass_balance}
\end{equation}
where $c_i$ is the average alignment cost for node $i$. This expression reveals a soft-thresholding behavior:
\begin{itemize}
    \item \textbf{Well-aligned nodes} ($c_i \ll \rho$): The exponential is near unity, so $\boldsymbol{r}_i^* \approx \mu_i$---mass flows fully, preserving beneficial complementarity.
    \item \textbf{Conflicting nodes} ($c_i \gg \rho$): The exponential decays toward zero, suppressing $\boldsymbol{r}_i^*$ and effectively rejecting noisy correspondences.
\end{itemize}

The upshot is that OptiMAG can separate useful cross-modal complementarity from destructive noise, all without requiring explicit conflict labels. The coefficient $\rho$ controls how aggressively this filtering operates: smaller values reject misaligned nodes more readily (Noise Rejection in Figure~\ref{fig:framework}).

\subsubsection{The UFGW Objective}
Combining the above components, we arrive at the Unbalanced Fused Gromov-Wasserstein objective. Given a batch of $N$ nodes, we seek a transport plan $\boldsymbol{\pi} \in \mathbb{R}_+^{N \times N}$ that minimizes:
\begin{equation}
\begin{aligned}
    \mathcal{L}_{\text{UFGW}}(\boldsymbol{\pi}) = \quad & (1-\alpha) \sum_{i,j} \mathbf{M}_{ij} \boldsymbol{\pi}_{ij} \\
    + \quad & \alpha \sum_{i,j,k,l} L(\bar{\mathbf{C}}_m(i, k), \bar{\mathbf{C}}_G(j, l)) \boldsymbol{\pi}_{ij} \boldsymbol{\pi}_{kl} \\
    + \quad & \rho \left( \text{KL}(\boldsymbol{\pi}\mathbf{1}_N | \boldsymbol{\mu}_m) + \text{KL}(\boldsymbol{\pi}^\top\mathbf{1}_N | \boldsymbol{\nu}) \right) \\
    + \quad & \epsilon H(\boldsymbol{\pi})
\end{aligned}
\label{eq:ufgw_objective}
\end{equation}
The four terms correspond to anchoring, structural matching, marginal relaxation, and entropic smoothing, respectively. The entropic term $H(\boldsymbol{\pi}) = -\sum_{i,j} \boldsymbol{\pi}_{ij} \log \boldsymbol{\pi}_{ij}$ facilitates efficient Sinkhorn optimization. Hyperparameters $\alpha$, $\rho$, and $\epsilon$ control the relative emphasis on each component.

\subsection{Optimization and Training}
\label{subsec:optimization}

Direct optimization of Eq.~(\ref{eq:ufgw_objective}) is impractical for large graphs: the GW term involves $O(N^3)$ tensor operations. Two strategies make training feasible---subgraph sampling to reduce problem size, and entropic Sinkhorn iterations to enable efficient GPU computation.

\subsubsection{Efficient Subgraph Sinkhorn}
We train on mini-batches of $B$ nodes sampled via neighbor sampling. Although each batch is local, the graph cost matrix $\bar{\mathbf{C}}_G$ is extracted from the precomputed global PPR matrix (\cref{subsec:space_construction}), so structural context beyond the sampled subgraph is retained.

The UFGW objective is non-convex due to the quadratic GW term. We tackle it with Block Coordinate Descent (BCD): at each outer iteration, fixing $\boldsymbol{\pi}$ linearizes the quadratic term, reducing the problem to a KL-regularized linear transport that admits a closed-form Sinkhorn solution~\cite{sejourne2019sinkhorn}.

Concretely, let $\mathbf{G}^{(k)} = (1-\alpha)\mathbf{M} + 2\alpha \bar{\mathbf{C}}_m \boldsymbol{\pi}^{(k)} \bar{\mathbf{C}}_G^\top$ be the linearized gradient at BCD iteration $k$. Defining the Gibbs kernel $\mathbf{K} = \exp(-\mathbf{G}^{(k)} / \epsilon)$, the unbalanced Sinkhorn updates read:
\begin{equation}
    \mathbf{u} \leftarrow \left( \frac{\boldsymbol{\mu}_m}{\mathbf{K} \mathbf{v}} \right)^{\frac{\rho}{\rho + \epsilon}}, \quad \mathbf{v} \leftarrow \left( \frac{\boldsymbol{\nu}}{\mathbf{K}^\top \mathbf{u}} \right)^{\frac{\rho}{\rho + \epsilon}}
\end{equation}
The exponent $\frac{\rho}{\rho+\epsilon}$ arises from the unbalanced formulation and interpolates between entropy-regularized OT ($\rho \to \infty$) and fully unbalanced transport ($\rho \to 0$). After convergence, the transport plan is recovered as $\boldsymbol{\pi}^* = \text{diag}(\mathbf{u}) \mathbf{K} \text{diag}(\mathbf{v})$. Because the inner loop consists only of matrix-vector products, it parallelizes efficiently on modern GPUs.

\subsubsection{Overall Objective}
Once we obtain the optimal plan $\boldsymbol{\pi}^*$, the regularization loss is simply $\mathcal{L}_{\text{Reg}}^{(m)} = \mathcal{L}_{\text{UFGW}}(\boldsymbol{\pi}^*)$ for each modality $m$. We apply this regularizer to both text and image modalities independently, yielding the overall training objective:
\begin{equation}
    \mathcal{L}_{\text{Total}} = \mathcal{L}_{\text{Task}} + \lambda \sum_{m \in \{T, I\}} \mathcal{L}_{\text{Reg}}^{(m)}
\end{equation}
Here $\mathcal{L}_{\text{Task}}$ is the downstream supervision (e.g., cross-entropy for classification), and $\lambda$ controls regularization strength.

A subtle but important detail: we \texttt{detach} $\boldsymbol{\pi}^*$ during backpropagation. Gradients thus flow only through the cost matrix $\bar{\mathbf{C}}_m$ into the encoder, not through the transport plan itself. This design has two benefits: it forces the encoder to adapt its embeddings to align with the graph structure, and it ensures that nodes whose mass was truncated (i.e., those flagged as noisy by the KL mechanism) contribute no gradient signal.

\section{Experiments}
\label{sec:experiments}

To validate the effectiveness and robustness of OptiMAG, we establish a standardized experimental environment based on OpenMAG~\cite{openmag2026}. This section details our dataset selection, baseline configurations, and implementation specifics.

\subsection{Experimental Setup}
\label{subsec:setup}

\textbf{Datasets.} 
We draw six datasets from OpenMAG to stress-test generalization across varied graph topologies and modality configurations. \textbf{Graph-centric} tasks (node classification, link prediction, clustering) employ the Amazon co-purchase networks---\textbf{Grocery}, \textbf{Movies}, \textbf{Toys}---where edges capture product complementarity, plus \textbf{Reddit-S}, a strongly homophilous social-thread graph. \textbf{Multimodal-centric} generation (Graph-to-Text, Graph-to-Image) uses \textbf{Flickr30k} and \textbf{SemArt}; see Table~\ref{tab:datasets} for statistics.

\begin{table}[htbp]
\caption{Statistics of the datasets used in our experiments. Domain types include E-Commerce (E-Comm), Social Media (Social), Image Networks (Img), and Art Networks (Art).}
\label{tab:datasets}
\vskip 0.15in
\begin{center}
\begin{small}
\begin{sc}
\resizebox{\columnwidth}{!}{
\begin{tabular}{lcccc}
\toprule
\textbf{Dataset} & \textbf{Nodes} & \textbf{Edges} & \textbf{Modalities} & \textbf{Domain} \\
\midrule
Grocery & 17,074 & 171,340 & Text, Visual & E-Comm \\
Movies & 16,672 & 218,390 & Text, Visual & E-Comm \\
Toys & 20,695 & 126,886 & Text, Visual & E-Comm \\
Reddit-S & 15,894 & 566,160 & Text, Visual & Social \\
\midrule
Flickr30k & 31,783 & 181,151 & Text, Visual & Img \\
SemArt & 21,382 & 1,216,432 & Text, Visual & Art \\
\bottomrule
\end{tabular}
}
\end{sc}
\end{small}
\end{center}
\vskip -0.1in
\end{table}

\textbf{Baselines.}
\textbf{Unimodal GNNs} (\textbf{GCN}, \textbf{GAT}) operate on single-modality features or naive concatenation; they quantify the value of multimodal fusion. \textbf{Graph-enhanced models} (\textbf{DGF}~\cite{zhengDGF2025}, \textbf{DMGC}~\cite{guoDMGC2025}) address heterophily and structural noise---goals closely related to our conflict-resolution objective. \textbf{Multimodal graph models} (\textbf{MMGCN}~\cite{huMMGCNMultimodalFusion2021}, \textbf{MGAT}~\cite{taoMGATMultimodalGraph2020}, \textbf{UniGraph2}~\cite{heUniGraph2LearningUnified2025}) represent current best practices for MAG learning. All three multimodal baselines serve as backbones for OptiMAG integration; we report metrics both with and without the regularizer.

\textbf{Implementation Details.}
Experiments run on an NVIDIA RTX 6000 Ada (96\,GB) under PyTorch. Backbone hyperparameters (learning rate, depth, dropout) follow OpenMAG's recommended defaults~\cite{openmag2026} without modification. OptiMAG-specific hyperparameters are grid-searched: $\lambda \in \{10^{-3}, \ldots, 1\}$, $\tau \in \{0.1, 0.5, 1\}$, $\rho \in \{0.01, 0.1, 1\}$; Sinkhorn runs for 20 iterations. Graph diffusion uses PPR with teleport $\beta = 0.15$; all metrics are averaged over three runs.

\subsection{Performance on Graph-centric Tasks}
\label{subsec:graph_centric_results}

Table~\ref{tab:graph_centric_optimag} reports node classification, link prediction, and clustering results for MMGCN, MGAT, and UniGraph2---with and without OptiMAG---alongside DGF and LGMRec as reference baselines.

\begin{table*}[t]
\caption{Performance comparison on Graph-centric tasks (partial results shown). \textbf{Improv.} indicates the relative improvement over the original backbone. The best results within each backbone group are marked in \textbf{bold}.}
\label{tab:graph_centric_optimag}
\vskip 0.15in
\begin{center}
\resizebox{\textwidth}{!}{
\begin{tabular}{lcccccccc}
\toprule
\multirow{2}{*}{\textbf{Methods}} & \multicolumn{2}{c}{\textbf{Node Classification} (Movies)} & \multicolumn{2}{c}{\textbf{Node Classification} (RedditS)} & \multicolumn{2}{c}{\textbf{Link Prediction} (Toys)} & \multicolumn{2}{c}{\textbf{Node Clustering} (RedditS)} \\
\cmidrule(lr){2-3} \cmidrule(lr){4-5} \cmidrule(lr){6-7} \cmidrule(lr){8-9}
 & Acc & F1-score & Acc & F1-score & MRR & Hits@3 & NMI & ARI \\
\midrule
\multicolumn{9}{c}{\textit{Baselines}} \\
GCN    & 52.84 & 44.78 & 67.51 & 62.16 & 44.82 & 27.04 & 78.39 & 71.21 \\
GAT    & 51.38 & 43.22 & 67.12 & 62.44 & 45.23 & 27.51 & 78.40 & 68.90 \\
DGF    & 53.89 & 41.45 & 70.15 & 62.30 & 49.29 & 34.24 & 84.89 & 78.07 \\
LGMRec & 52.59 & 46.26 & 69.88 & 61.50 & 47.34 & 32.31 & 80.73 & 73.10 \\
\midrule
\multicolumn{9}{c}{\textit{Backbones w/ and w/o OptiMAG}} \\
MMGCN  & 52.12 & 41.85 & 68.45 & 60.20 & 45.39 & 27.56 & 67.90 & 50.42 \\
\textbf{+ OptiMAG} & \textbf{53.05} \small{(+0.93)} & \textbf{43.10} \small{(+1.25)} & \textbf{70.12} \small{(+1.67)} & \textbf{61.85} \small{(+1.65)} & \textbf{46.85} \small{(+1.46)} & \textbf{29.10} \small{(+1.54)} & \textbf{70.45} \small{(+2.55)} & \textbf{53.80} \small{(+3.38)} \\
\midrule
MGAT   & 50.00 & 37.31 & 66.80 & 58.90 & 46.72 & 30.02 & 73.36 & 60.55 \\
\textbf{+ OptiMAG} & \textbf{51.25} \small{(+1.25)} & \textbf{38.95} \small{(+1.64)} & \textbf{68.50} \small{(+1.70)} & \textbf{60.45} \small{(+1.55)} & \textbf{47.90} \small{(+1.18)} & \textbf{31.55} \small{(+1.53)} & \textbf{75.20} \small{(+1.84)} & \textbf{63.10} \small{(+2.55)} \\
\midrule
UniGraph2 & 46.78 & 31.03 & 60.50 & 52.15 & 10.34 & 3.09 & 31.85 & 12.85 \\
\textbf{+ OptiMAG} & \textbf{49.88} \small{(+3.10)} & \textbf{34.20} \small{(+3.17)} & \textbf{65.10} \small{(+4.60)} & \textbf{56.80} \small{(+4.65)} & \textbf{14.50} \small{(+4.16)} & \textbf{5.85} \small{(+2.76)} & \textbf{36.40} \small{(+4.55)} & \textbf{16.90} \small{(+4.05)} \\
\bottomrule
\end{tabular}
}
\end{center}
\vskip -0.1in
\end{table*}

OptiMAG delivers consistent gains across all backbones, tasks, and datasets---approximately +2\% accuracy for node classification and +3\% NMI for clustering. The uniformity of these improvements validates the method's broad applicability.

Notably, UniGraph2---a large pretrained model that struggles to adapt to specific graph structures out of the box---benefits most from OptiMAG, gaining up to +4.6\% accuracy on Reddit-S. This suggests that OptiMAG's structured gradients steer pretrained encoders toward finer-grained graph alignment.

\subsection{Performance on Multimodal-centric Tasks}
\label{subsec:mm_centric_results}

We further evaluate OptiMAG on generation tasks: \textbf{Graph-to-Text} (Flickr30k) and \textbf{Graph-to-Image} (SemArt). Table~\ref{tab:mm_generation_optimag} summarizes the results.

\begin{table}[htbp]
\caption{Performance comparison on Multimodal Generation tasks. We report BLEU-4 and CIDEr for G2Text, and CLIP-Score (CLIP-S) and DINOv2-Score (DINO-S) for G2Image. \textbf{Improv.} indicates relative gains.}
\label{tab:mm_generation_optimag}
\vskip 0.15in
\begin{center}
\resizebox{\columnwidth}{!}{
\begin{tabular}{lcccc}
\toprule
\multirow{2}{*}{\textbf{Methods}} & \multicolumn{2}{c}{\textbf{G2Text} (Flickr30k)} & \multicolumn{2}{c}{\textbf{G2Image} (SemArt)} \\
\cmidrule(lr){2-3} \cmidrule(lr){4-5}
 & BLEU-4 & CIDEr & CLIP-S & DINO-S \\
\midrule
\multicolumn{5}{c}{\textit{Baselines}} \\
GCN    & 5.69 & 38.44 & 67.15 & 49.65 \\
GAT    & 5.80 & 39.10 & 67.40 & 50.20 \\
DGF    & 6.83 & 44.28 & 68.43 & 52.30 \\
LGMRec & 5.95 & 39.00 & 68.47 & 52.73 \\
\midrule
\multicolumn{5}{c}{\textit{Backbones w/ and w/o OptiMAG}} \\
MMGCN  & 5.72 & 38.50 & 67.20 & 50.15 \\
\textbf{+ OptiMAG} & \textbf{6.55} \small{(+0.83)} & \textbf{43.10} \small{(+4.60)} & \textbf{68.95} \small{(+1.75)} & \textbf{52.88} \small{(+2.73)} \\
\midrule
MGAT   & 5.80 & 39.12 & 67.55 & 50.80 \\
\textbf{+ OptiMAG} & \textbf{6.62} \small{(+0.82)} & \textbf{43.55} \small{(+4.43)} & \textbf{69.10} \small{(+1.55)} & \textbf{53.05} \small{(+2.25)} \\
\midrule
UniGraph2 & 7.15 & 45.60 & 69.50 & 53.40 \\
\textbf{+ OptiMAG} & \textbf{7.48} \small{(+0.33)} & \textbf{47.85} \small{(+2.25)} & \textbf{70.82} \small{(+1.32)} & \textbf{54.65} \small{(+1.25)} \\
\bottomrule
\end{tabular}
}
\end{center}
\vskip -0.1in
\end{table}

OptiMAG again yields consistent improvements. The +4.6 CIDEr gain for MMGCN on G2Text is particularly notable: because CIDEr measures semantic alignment between generated and reference captions, this jump indicates that OptiMAG helps encoders capture more accurate entity information from the graph context.

\subsection{Ablation Study}
\label{subsec:ablation}

We ablate OptiMAG's components on Reddit-S (node classification) and Flickr30k (G2Text) using UniGraph2. Three variants are examined: \textbf{w/o UOT} removes KL relaxation ($\rho \!\to\! \infty$), enforcing strict marginals; \textbf{w/o GW} drops the structural term ($\alpha\!=\!0$), reducing OptiMAG to feature-only alignment; \textbf{w/o Prior} removes the anchor matrix $\mathbf{M}$, relying purely on learned correspondences.

\begin{table}[htbp]
\caption{Ablation study on key components of OptiMAG using UniGraph2 backbone. \textbf{Acc} and \textbf{F1} are reported for Reddit-S; \textbf{BLEU-4} (B-4) and \textbf{CIDEr} (C) for Flickr30k. The \textbf{Full Model} results align with Table \ref{tab:graph_centric_optimag} and Table \ref{tab:mm_generation_optimag}.}
\label{tab:ablation}
\vskip 0.15in
\begin{center}
\resizebox{0.95\columnwidth}{!}{
\begin{tabular}{lcccc}
\toprule
\multirow{2}{*}{\textbf{Variants}} & \multicolumn{2}{c}{\textbf{Reddit-S} (Node Classification)} & \multicolumn{2}{c}{\textbf{Flickr30k} (G2Text)} \\
\cmidrule(lr){2-3} \cmidrule(lr){4-5}
    & Acc & F1 & B-4 & C \\
\midrule
\textbf{Full Model} & \textbf{65.10} & \textbf{56.80} & \textbf{7.48} & \textbf{47.85} \\
\midrule
w/o UOT (Balanced) & 61.80 \small{(-3.30)} & 53.50 \small{(-3.30)} & 7.30 \small{(-0.18)} & 46.90 \small{(-0.95)} \\
w/o GW (Structure-free) & 61.00 \small{(-4.10)} & 52.60 \small{(-4.20)} & 7.05 \small{(-0.43)} & 45.30 \small{(-2.55)} \\
w/o Prior (Anchor-free) & 63.50 \small{(-1.60)} & 55.20 \small{(-1.60)} & 7.40 \small{(-0.08)} & 47.20 \small{(-0.65)} \\
\bottomrule
\end{tabular}
}
\end{center}
\vskip -0.1in
\end{table}

\textbf{Impact of Mass Relaxation (UOT).}
Removing UOT drops Reddit-S accuracy by 3.3\%. Reddit-S contains many heterophilous edges from casual replies; under strict marginal constraints the optimizer is forced to align noisy nodes, polluting encoder gradients. KL-based mass relaxation lets the model ``cut off'' such nodes adaptively.

\textbf{Impact of Structural Alignment (GW).}
Dropping the GW term causes the steepest decline (CIDEr $-$2.55 on Flickr30k). Without it, OptiMAG reduces to point-wise feature matching, losing the ability to preserve higher-order relational structure---confirming that explicit topology alignment is essential for cross-modal fusion.

\textbf{Impact of Prior Anchor (Prior).}
Removing the anchor matrix yields a smaller but noticeable drop. Although UFGW can learn correspondences from scratch, seeding it with the identity prior accelerates convergence and prevents semantically plausible but identity-incorrect misalignments.

\subsection{Sensitivity and Efficiency Analysis}
\label{subsec:sensitivity_efficiency}

We conclude with hyperparameter sensitivity and scalability analyses.

\subsubsection{Hyperparameter Sensitivity}
\begin{figure}[htbp]
    \centering
    \includegraphics[width=0.48\textwidth]{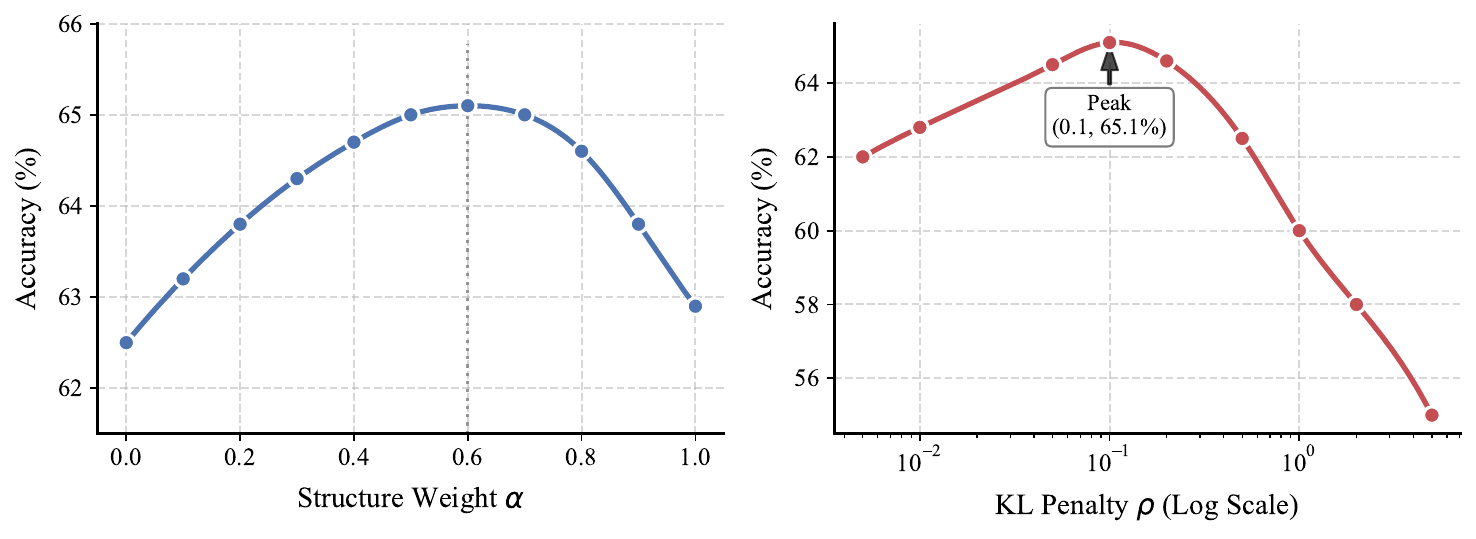}
    \caption{Hyperparameter sensitivity analysis on Reddit-S (Node Classification). \textbf{Left}: Impact of structure weight $\alpha$, which balances the prior anchor (Wasserstein term) and structural alignment (GW term). Performance peaks at $\alpha \approx 0.6$. \textbf{Right}: Impact of KL penalty $\rho$, which controls tolerance to marginal deviation. An inverted U-shape is observed, with the optimum at $\rho \approx 0.1$.}
    \label{fig:sensitivity}
\end{figure}

Figure~\ref{fig:sensitivity} shows how structural weight $\alpha$ and KL penalty $\rho$ affect node classification on Reddit-S (UniGraph2 backbone).

\textbf{Balance between Feature and Structure ($\alpha$).}
Accuracy improves as $\alpha$ rises from 0 to 0.6, confirming the value of structural alignment; beyond 0.6, over-reliance on topology at the expense of identity priors causes alignment drift. The sweet spot lies in $\alpha \in [0.5, 0.7]$.

\textbf{Tolerance to Structural Noise ($\rho$).}
An inverted-U response to $\rho$ is observed: too small a value discards useful mass, while too large a value reverts to balanced OT. Peak performance at $\rho \approx 0.1$ indicates that moderate relaxation best filters heterophilous noise.

\subsubsection{Scalability and Efficiency}
\begin{figure}[htbp]
\centering
\includegraphics[width=0.48\textwidth]{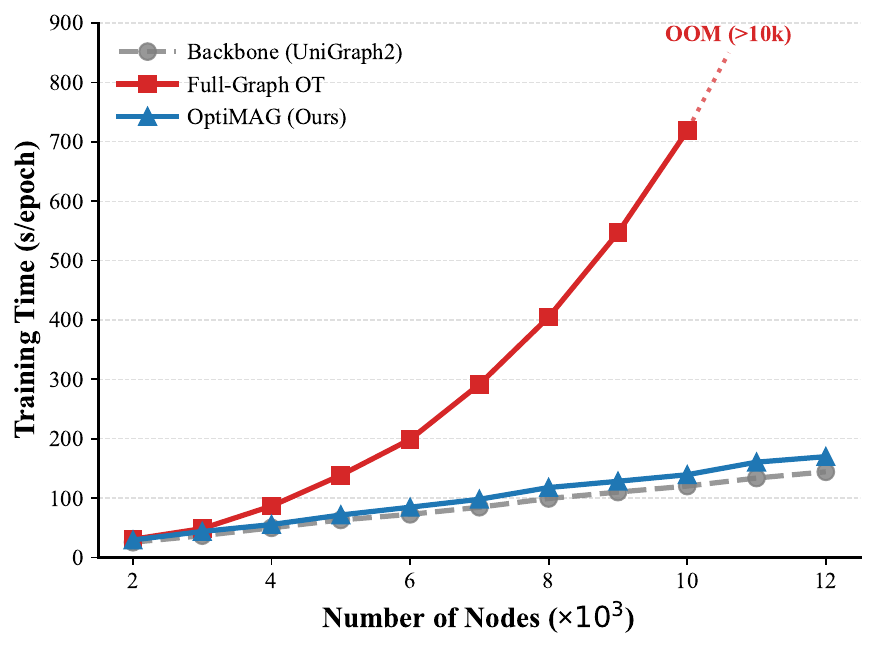}
\caption{Scalability analysis on Reddit-S. We report the training time per epoch w.r.t. the number of graph nodes. OptiMAG maintains linear scalability, whereas Full-Graph OT exhibits cubic complexity and triggers OOM (Out Of Memory) errors when nodes $>10k$.}
\label{fig:efficiency}
\end{figure}

GW computation is $O(N^3)$, prohibitive for full graphs. OptiMAG sidesteps this via subgraph-sampled Sinkhorn. Figure~\ref{fig:efficiency} compares per-epoch training times on Reddit-S subsets (2k--12k nodes) for UniGraph2, Full-Graph OT, and OptiMAG (batch size 512).

Full-Graph OT hits OOM beyond 10k nodes; OptiMAG scales linearly because OT is confined to fixed-size batches. The regularizer adds only 15--20\% overhead, making it practical for million-node MAG tasks.

\section{Related Work}
\label{sec:related_work}

\textbf{Multimodal Graph Learning.}
Standardized MAG benchmarks \cite{yanWhenGraphMeets2024, zhuMosaicModalitiesComprehensive2025} have driven rapid progress. Early methods \cite{huMMGCNMultimodalFusion2021, taoMGATMultimodalGraph2020} project modality features into a shared space before GNN aggregation; recent foundation-model approaches like UniGraph2 \cite{heUniGraph2LearningUnified2025} and GRAPHGPT-O \cite{fangGRAPHGPTOSynergisticMultimodal2025} scale this paradigm further. Yet all propagate over a fixed explicit graph, overlooking structural-semantic conflicts---edges that are homophilous in one modality but heterophilous in another. We address this with an OT-based regularizer enforcing local cross-modal alignment.

\textbf{Optimal Transport for Alignment.}
OT offers a principled framework for comparing distributions \cite{peyre2019computational}, made practical by entropic Sinkhorn acceleration \cite{cuturi2013sinkhorn}. The Gromov-Wasserstein (GW) distance \cite{memoli2011gromov} matches relational structure across incomparable spaces; Vayer et al.\ \cite{vayer2019optimal} extend it to the Fused GW (FGW) formulation that jointly considers features and structure. To handle noisy correspondences, we adopt unbalanced OT \cite{sejourne2019sinkhorn}, relaxing marginal constraints via KL divergence. Unlike prior global-alignment OT methods \cite{xu2019gromov, caoOTKGEMultimodalKnowledge2022}, OptiMAG operates at the \textit{local neighborhood} scale to match GNN receptive fields.

\section{Conclusion}
\label{sec:conclusion}

We identify structural-semantic conflict in MAGs---edges that are homophilous in one modality yet heterophilous in another---and show that standard GNN aggregation propagates this inconsistency, injecting modality-specific noise into learned representations. OptiMAG mitigates the problem via Unbalanced Optimal Transport: Fused Gromov-Wasserstein aligns each modality's implicit semantic structure with the explicit graph, while KL-relaxed marginals let the transport plan reject severely conflicting nodes rather than forcing spurious correspondences. Subgraph-sampled Sinkhorn iterations reduce GW's $O(N^3)$ cost to linear overhead, enabling million-node deployment.

Across six benchmarks and three backbones, OptiMAG yields consistent improvements in both graph-centric (node classification, link prediction, clustering) and generation (graph-to-text, graph-to-image) tasks; gains are largest for pretrained encoders like UniGraph2, where the structured regularization steers large models toward finer-grained graph adaptation. Ablation studies confirm that both FGW-based structural alignment and UOT-based noise rejection are essential. Future work may extend the framework to additional modalities (audio, point clouds) and dynamic graphs.

\section*{Impact Statement}

This paper presents work whose goal is to advance the field of Machine
Learning. There are many potential societal consequences of our work, none
which we feel must be specifically highlighted here.

\bibliography{ot}

\begin{thebibliography}{25}
\providecommand{\natexlab}[1]{#1}
\providecommand{\url}[1]{\texttt{#1}}
\expandafter\ifx\csname urlstyle\endcsname\relax
  \providecommand{\doi}[1]{doi: #1}\else
  \providecommand{\doi}{doi: \begingroup \urlstyle{rm}\Url}\fi

\bibitem[Authors(2026)]{openmag2026}
Authors, A.
\newblock Openmag: A comprehensive benchmark for multimodal-attributed graph.
\newblock Under Review at ICML 2026, 2026.
\newblock Available at \url{https://anonymous.4open.science/r/OpenMAG-F703/}.

\bibitem[Cai et~al.(2025)Cai, Liu, Gao, Jiang, Zhang, van~den Hengel, and Shi]{caiValueCrossModalMisalignment2025}
Cai, Y., Liu, Y., Gao, E., Jiang, T., Zhang, Z., van~den Hengel, A., and Shi, J.~Q.
\newblock On the {{Value}} of {{Cross-Modal Misalignment}} in {{Multimodal Representation Learning}}, September 2025.

\bibitem[Cao et~al.(2022)Cao, Xu, Yang, He, Cao, and Huang]{caoOTKGEMultimodalKnowledge2022}
Cao, Z., Xu, Q., Yang, Z., He, Y., Cao, X., and Huang, Q.
\newblock \textbraceleft{{OTKGE}}:\textbraceright{} {{Multi-modal Knowledge Graph Embeddings}} via {{Optimal Transport}}.
\newblock In \emph{Proceedings of the {{Advances}} in {{Neural Information Processing Systems}} 35: {{Annual Conference}} on {{Neural Information Processing Systems}} 2022}, 2022.

\bibitem[Chizat et~al.(2018)Chizat, Peyr{\'e}, Schmitzer, and Vialard]{chizat2018scaling}
Chizat, L., Peyr{\'e}, G., Schmitzer, B., and Vialard, F.-X.
\newblock Scaling algorithms for unbalanced optimal transport problems.
\newblock \emph{Mathematics of Computation}, 87\penalty0 (314):\penalty0 2563--2609, 2018.

\bibitem[Cuturi(2013)]{cuturi2013sinkhorn}
Cuturi, M.
\newblock Sinkhorn distances: Lightspeed computation of optimal transport.
\newblock In \emph{Advances in Neural Information Processing Systems (NeurIPS)}, volume~26, pp.\  2292--2300, 2013.

\bibitem[Fang et~al.(2025)Fang, Jin, Shen, Ding, Tan, and Han]{fangGRAPHGPTOSynergisticMultimodal2025}
Fang, Y., Jin, B., Shen, J., Ding, S., Tan, Q., and Han, J.
\newblock \textbraceleft{{GRAPHGPT-O}}:\textbraceright{} {{Synergistic Multimodal Comprehension}} and {{Generation}} on {{Graphs}}.
\newblock In \emph{Proceedings of the \textbraceleft{{IEEE}}/{{CVF}}\textbraceright{} {{Conference}} on {{Computer Vision}} and {{Pattern Recognition}}}, volume abs/2502.11925, pp.\  19467--19476. Computer Vision Foundation / \textbraceleft IEEE\textbraceright, 2025.
\newblock \doi{10.1109/CVPR52734.2025.01813}.

\bibitem[Gasteiger et~al.(2019)Gasteiger, Bojchevski, and G{\"u}nnemann]{gasteiger2019predict}
Gasteiger, J., Bojchevski, A., and G{\"u}nnemann, S.
\newblock Predict then propagate: Graph neural networks meet personalized pagerank.
\newblock In \emph{International Conference on Learning Representations (ICLR)}, 2019.

\bibitem[He et~al.(2025)He, Sui, He, Liu, Sun, and Hooi]{heUniGraph2LearningUnified2025}
He, Y., Sui, Y., He, X., Liu, Y., Sun, Y., and Hooi, B.
\newblock {{UniGraph2}}: {{Learning}} a {{Unified Embedding Space}} to {{Bind Multimodal Graphs}}.
\newblock In \emph{Proceedings of the \textbraceleft{{ACM}}\textbraceright{} on {{Web Conference}} 2025}, pp.\  1759--1770. \textbraceleft ACM\textbraceright, 2025.
\newblock \doi{10.1145/3696410.3714818}.

\bibitem[Hu et~al.(2021)Hu, Liu, Zhao, and Jin]{huMMGCNMultimodalFusion2021}
Hu, J., Liu, Y., Zhao, J., and Jin, Q.
\newblock \textbraceleft{{MMGCN}}:\textbraceright{} {{Multimodal Fusion}} via {{Deep Graph Convolution Network}} for {{Emotion Recognition}} in {{Conversation}}.
\newblock In \emph{Proceedings of the 59th {{Annual Meeting}} of the {{Association}} for {{Computational Linguistics}} and the 11th {{International Joint Conference}} on {{Natural Language Processing}}}, pp.\  5666--5675. Association for Computational Linguistics, 2021.
\newblock \doi{10.18653/V1/2021.ACL-LONG.440}.

\bibitem[Li et~al.(2024{\natexlab{a}})]{liLearnHeterophily2024}
Li, H. et~al.
\newblock Learn from heterophily: Heterophilous information-enhanced graph neural network.
\newblock \emph{arXiv preprint arXiv:2403.07540}, 2024{\natexlab{a}}.

\bibitem[Li et~al.(2024{\natexlab{b}})]{liDeepHeterophilyGraph2024}
Li, K. et~al.
\newblock Deep heterophily graph rewiring.
\newblock In \emph{Proceedings of the Web Conference 2024 (WWW)}, 2024{\natexlab{b}}.

\bibitem[M{\'e}moli(2011)]{memoli2011gromov}
M{\'e}moli, F.
\newblock Gromov-wasserstein distances and the metric approach to object matching.
\newblock In \emph{Foundations of Computational Mathematics}, 2011.

\bibitem[Peyr{\'e} \& Cuturi(2019)Peyr{\'e} and Cuturi]{peyre2019computational}
Peyr{\'e}, G. and Cuturi, M.
\newblock Computational optimal transport.
\newblock \emph{Foundations and Trends{\textregistered} in Machine Learning}, 11\penalty0 (5-6):\penalty0 355--607, 2019.

\bibitem[Reimers \& Gurevych(2019)Reimers and Gurevych]{reimers2019sentence}
Reimers, N. and Gurevych, I.
\newblock Sentence-bert: Sentence embeddings using siamese bert-networks.
\newblock In \emph{Proceedings of the 2019 Conference on Empirical Methods in Natural Language Processing (EMNLP)}, pp.\  3982--3992, 2019.

\bibitem[S{\'e}journ{\'e} et~al.(2019)S{\'e}journ{\'e}, Feydy, Vialard, Trouv{\'e}, and Peyr{\'e}]{sejourne2019sinkhorn}
S{\'e}journ{\'e}, T., Feydy, J., Vialard, F.-X., Trouv{\'e}, A., and Peyr{\'e}, G.
\newblock Sinkhorn divergences for unbalanced optimal transport.
\newblock \emph{arXiv preprint arXiv:1910.12958}, 2019.

\bibitem[Tao et~al.(2020)Tao, Wei, Wang, He, Huang, and Chua]{taoMGATMultimodalGraph2020}
Tao, Z., Wei, Y., Wang, X., He, X., Huang, X., and Chua, T.-{\textbraceright}.
\newblock \textbraceleft{{MGAT}}:\textbraceright{} {{Multimodal Graph Attention Network}} for {{Recommendation}}.
\newblock \emph{Inf. Process. Manag.}, 57\penalty0 (5):\penalty0 102277, 2020.
\newblock ISSN 0306-4573.
\newblock \doi{10.1016/J.IPM.2020.102277}.

\bibitem[Vayer et~al.(2019{\natexlab{a}})Vayer, Chapel, Flamary, Tavenard, and Courty]{vayer2019fused}
Vayer, T., Chapel, L., Flamary, R., Tavenard, R., and Courty, N.
\newblock Fused gromov-wasserstein distance for structured objects.
\newblock \emph{Algorithms}, 13\penalty0 (9):\penalty0 212, 2019{\natexlab{a}}.

\bibitem[Vayer et~al.(2019{\natexlab{b}})Vayer, Courty, Tavenard, Chapel, and Flamary]{vayer2019optimal}
Vayer, T., Courty, N., Tavenard, R., Chapel, L., and Flamary, R.
\newblock Optimal transport for structured data with application on graphs.
\newblock In \emph{International Conference on Machine Learning (ICML)}, pp.\  6275--6284, 2019{\natexlab{b}}.

\bibitem[Wang et~al.(2025)Wang, Xu, Zhang, Feng, and Gao]{guoDMGC2025}
Wang, Q., Xu, H., Zhang, Z., Feng, W., and Gao, Q.
\newblock Deep multi-modal graph clustering via graph transformer network.
\newblock In \emph{Proceedings of the AAAI Conference on Artificial Intelligence}, volume~39, pp.\  7835--7843, 2025.
\newblock \doi{10.1609/aaai.v39i8.32844}.

\bibitem[Xu et~al.(2019)]{xu2019gromov}
Xu, H. et~al.
\newblock Gromov-wasserstein learning for graph matching and node embedding.
\newblock In \emph{International Conference on Machine Learning (ICML)}, pp.\  6932--6941, 2019.

\bibitem[Yan et~al.(2024)Yan, Li, Yin, Yu, Han, Li, Zeng, Sun, and Wang]{yanWhenGraphMeets2024}
Yan, H., Li, C., Yin, J., Yu, Z., Han, W., Li, M., Zeng, Z., Sun, H., and Wang, S.
\newblock When {{Graph}} meets {{Multimodal}}: {{Benchmarking}} and {{Meditating}} on {{Multimodal Attributed Graphs Learning}}.
\newblock 2024.
\newblock \doi{10.48550/arXiv.2410.09132}.

\bibitem[Zheng et~al.(2026)Zheng, Wang, Xu, and Yang]{zhengDGF2025}
Zheng, H., Wang, H., Xu, J., and Yang, R.
\newblock Cross-contrastive clustering for multimodal attributed graphs with dual graph filtering.
\newblock In \emph{Proceedings of the 32nd ACM SIGKDD Conference on Knowledge Discovery and Data Mining}, 2026.
\newblock To appear.

\bibitem[Zheng et~al.(2024)]{zhengGNNHeterophilySurvey2024}
Zheng, Y. et~al.
\newblock Graph neural networks for graphs with heterophily: A survey.
\newblock \emph{arXiv preprint arXiv:2402.05085}, 2024.

\bibitem[Zhu et~al.(2020)Zhu, Yan, Zhao, Heimann, Akoglu, and Koutra]{zhuBeyondHomophilyGraph2020}
Zhu, J., Yan, Y., Zhao, L., Heimann, M., Akoglu, L., and Koutra, D.
\newblock Beyond homophily in graph neural networks: Current limitations and effective designs.
\newblock In \emph{Advances in Neural Information Processing Systems (NeurIPS)}, volume~33, pp.\  7793--7804, 2020.

\bibitem[Zhu et~al.(2025)Zhu, Zhou, Qian, He, Zhao, Shah, and Koutra]{zhuMosaicModalitiesComprehensive2025}
Zhu, J., Zhou, Y., Qian, S., He, Z., Zhao, T., Shah, N., and Koutra, D.
\newblock Mosaic of {{Modalities}}: \textbraceleft{{A}}\textbraceright{} {{Comprehensive Benchmark}} for {{Multimodal Graph Learning}}.
\newblock In \emph{Proceedings of the \textbraceleft{{IEEE}}/{{CVF}}\textbraceright{} {{Conference}} on {{Computer Vision}} and {{Pattern Recognition}}}, pp.\  14215--14224. Computer Vision Foundation / \textbraceleft IEEE\textbraceright, 2025.
\newblock \doi{10.1109/CVPR52734.2025.01326}.

\end{thebibliography}
\bibliographystyle{icml2026}

\newpage
\appendix
\onecolumn

\section{Notation Reference}
\label{appendix:notation}

\begin{table}[htbp]
\caption{Summary of key notation used in this paper.}
\label{tab:notation}
\vskip 0.1in
\begin{center}
\begin{tabular}{@{}clcl@{}}
\toprule
\textbf{Symbol} & \textbf{Description} & \textbf{Symbol} & \textbf{Description} \\
\midrule
$\mathcal{G}$ & Multimodal Attributed Graph & $\boldsymbol{\pi}$ & Transport plan (coupling) \\
$\mathcal{V}, \mathcal{E}$ & Node set, edge set & $\mathcal{W}, \mathcal{GW}$ & Wasserstein, Gromov-W.\ distance \\
$N$ & Number of nodes & $\mathbf{C}_m$ & Modality cost matrix \\
$\mathcal{M}$ & Modality set $\{T, I\}$ & $\bar{\mathbf{C}}_m$ & Normalized modality cost \\
$\mathbf{A}$ & Adjacency matrix & $\mathbf{C}_G$ & Graph cost matrix (PPR-based) \\
$\mathbf{h}_i^m$ & Embedding of node $i$, modality $m$ & $\bar{\mathbf{C}}_G$ & Normalized graph cost \\
$\mathbf{H}^m$ & Batch embedding matrix & $\mathbf{M}_{ij}$ & Anchor prior matrix \\
$f_{\theta_m}$ & Modality-specific encoder & $\mathbf{K}$ & Gibbs kernel $\exp(-\mathbf{G}/\epsilon)$ \\
$\boldsymbol{\mu}, \boldsymbol{\nu}$ & Source/target distributions & $\mathbf{u}, \mathbf{v}$ & Sinkhorn scaling vectors \\
\midrule
$\mathcal{L}_{\text{UFGW}}$ & UFGW objective & $\alpha$ & Structure vs.\ anchor weight \\
$\mathcal{L}_{\text{Reg}}$ & Regularization loss & $\rho$ & KL penalty coefficient \\
$\mathcal{L}_{\text{Task}}$ & Task-specific loss & $\tau$ & Anchor threshold \\
$\mathcal{L}_{\text{Total}}$ & Total training objective & $\lambda$ & Regularization strength \\
\bottomrule
\end{tabular}
\end{center}
\vskip -0.1in
\end{table}

\section{Empirical Analysis}
\label{appendix:empirical_analysis}

Through this empirical analysis, we aim to demonstrate that in real-world MAGs, there exist edges where the two connected nodes exhibit high similarity in only some modalities, rather than in all modalities.

We first selected 10 MAG datasets, then used the Qwen-2-VL-7B multimodal large language model (MLLM) to encode the text and images in each dataset. Subsequently, we randomly sampled 1000 edges from each dataset, computed the cosine similarity between the two connected nodes in both text and image modalities, and plotted them in scatter plots.

The following presents the empirical analysis results across the datasets:

\begin{figure}[htbp]
    \centering
    \includegraphics[width=1\textwidth]{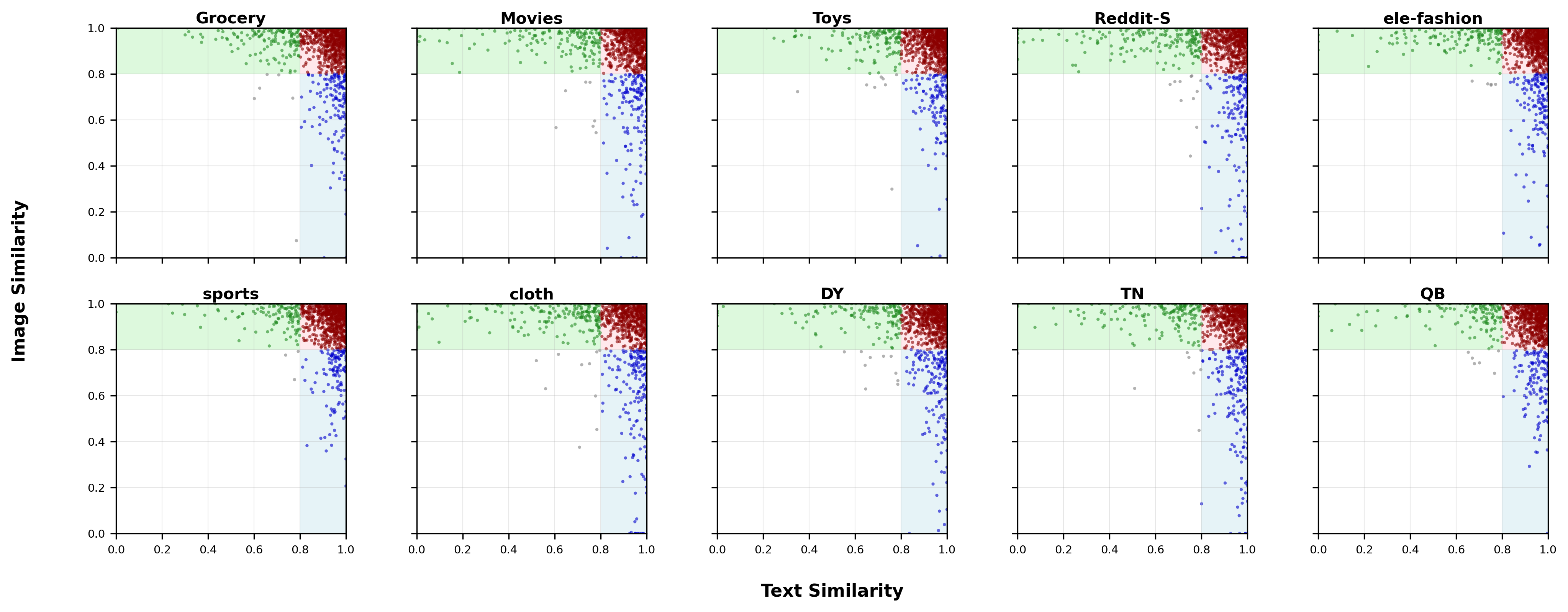}
    \caption{Empirical analysis across multiple MAG datasets. Each point represents an edge in the corresponding dataset, with 1000 edges sampled per dataset. The x-axis indicates the cosine similarity of text modality embeddings between two connected nodes, while the y-axis indicates the cosine similarity of image modality embeddings.}
    \label{fig:empirical_analysis}
\end{figure}

As shown in the figure above, there are substantial numbers of node pairs in the blue region (text similar but image dissimilar) and the green region (image similar but text dissimilar), comparable to the number of node pairs in the red region (both text and image similar). This demonstrates that in real-world MAGs, many node pairs exhibit high similarity in only some modalities rather than in all modalities.

\end{document}